\documentclass[journal, twoside]{IEEEtran}
\usepackage[english]{babel}

\usepackage[letterpaper,top=2.01cm,bottom=1.52cm,left=1.69cm,right=1.69cm,marginparwidth=1.75cm]{geometry}

\usepackage{amsfonts}
\usepackage{amsmath, bm}
\usepackage{graphicx}
\usepackage{algorithm}
\usepackage[mathscr]{euscript}
\usepackage[noend]{algpseudocode}
\usepackage[colorlinks=true, allcolors=blue]{hyperref}
\usepackage{verbatim} 
\usepackage{multirow}
\usepackage{booktabs}
\usepackage{cite}
\usepackage{caption}
\usepackage{subcaption}

\usepackage{scrextend}

\usepackage{algorithm}
\usepackage{algpseudocode}
\usepackage{multirow}
\usepackage{pifont}
\newcommand{\cmark}{\ding{51}}%
\newcommand{\xmark}{\ding{55}}%
\usepackage{graphicx}
\usepackage{subcaption}
\usepackage[export]{adjustbox}
\usepackage[x11names]{xcolor}
\usepackage{marvosym}

\begin{document}

\title{Multi-Echo Denoising in Adverse Weather}
\author{Alvari Seppänen$^{1}$, Risto Ojala$^{2}$, Kari Tammi$^{1}$

\thanks{$^{1}$A. Seppänen, and Kari Tammi are with Aalto University and University of Helsinki, Finland.
{\tt\footnotesize alvari.seppanen@aalto.fi}}%
\thanks{$^{2}$R. Ojala is with Aalto University, Finland.}}

\markboth{PREPRINT VERSION}%
{Seppänen \MakeLowercase{\textit{et al.}}: Multi-Echo Denoising in Adverse Weather}

\maketitle

\begin{abstract}
Adverse weather can cause noise to light detection and ranging (LiDAR) data.
This is a problem since it is used in many outdoor applications, \textit{e.g.} object detection and mapping. 
We propose the task of multi-echo denoising, where the goal is to pick the echo that represents the objects of interest and discard other echoes. 
Thus, the idea is to pick points from alternative echoes that are not available in standard strongest echo point clouds due to the noise.
In an intuitive sense, we are trying to see through the adverse weather.
To achieve this goal, we propose a novel self-supervised deep learning method and the characteristics similarity regularization method to boost its performance.
Based on extensive experiments on a semi-synthetic dataset, our method achieves superior performance compared to the state-of-the-art in self-supervised adverse weather denoising (23\% improvement). 
Moreover, the experiments with a real multi-echo adverse weather dataset prove the efficacy of multi-echo denoising.
Our work enables more reliable point cloud acquisition in adverse weather and thus promises safer autonomous driving and driving assistance systems in such conditions.
The code is available at https://github.com/alvariseppanen/SMEDNet
\end{abstract}

\begin{IEEEkeywords}
Deep Learning for Visual Perception, AI-Based Methods, Intelligent Transportation Systems, Visual Learning, Computer Vision for Transportation.
\end{IEEEkeywords}


\section{Introduction}
\label{sec:intro}

The impact of adverse weather conditions on light detection and ranging (LiDAR) sensor data can be enormous.
Airborne particles, including rain \cite{wallace2020full, goodin2019predicting, Bijelic_2020_STF}, snowfall \cite{jokela2019testing, kutila2020benchmarking, michaud2015towards, o1970visibility, Bijelic_2020_STF}, and fog \cite{bijelic2018benchmark, wallace2020full, heinzler2019weather, Bijelic_2020_STF} cause unwanted absorptions, refractions, and reflections of the LIDAR signal, which causes cluttered and missing points.
This is a critical issue as point clouds are typically used for determining the accessible volume of the environment, for instance in obstacle detection methods.
Furthermore, it affects other downstream perception algorithms, namely object detection \cite{do2022lossdistillnet, do2022missvoxelnet, hahner2022lidar, hahner2021fog}, which is a vital component, \textit{e.g.} in automated driving and driving assistance systems.
Moreover, accident rates for human drivers are notably higher in adverse weather conditions, as reported by the European Commission \cite{RoadsafetyintheEuropeanUnion} and the US Department of Transportation \cite{HowDoWeatherEventsImpactRoads}.
Therefore, reliable perception data is crucial in such conditions.

\begin{figure}[t]
\centering
\includegraphics[width=0.48\textwidth]{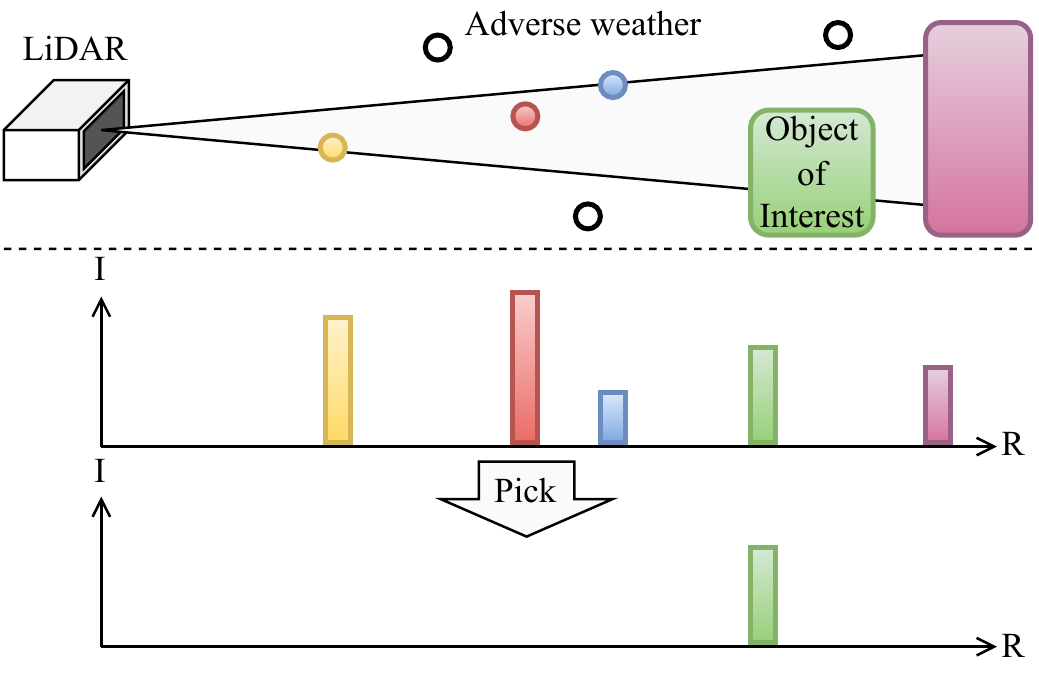}
\caption{In the concept of multi-echo denoising, multiple echoes are acquired for a single emitted pulse and the echo caused by the object of interest is picked. Generally in single-echo approaches, only the strongest echo (red) is available. Thus, crucial information about the object of interest can be lost. I and R denote intensity and range, respectively.}
\label{fig:idea}
\end{figure}

Previous LiDAR adverse weather denoising work has mainly used single-echo point clouds, where noise caused by adverse weather is removed from the point cloud.
This work proposes to use multi-echo point clouds and pick the echo that represents the objects of interest and discard the echoes that represent airborne particles (noise) or irrelevant artifacts caused by refractions, for example.
We use point cloud data format as it is provided by most of-the-shelf LiDARs, however, using raw photon count histograms has been shown to be beneficial when denoising LiDAR data in fog \cite{sang2022histogram}.

Our multi-echo approach is beneficial in several ways.
First, the approach utilizes information that is unavailable in single-echo approaches.
That is, objects might be occluded by airborne particles in single-echo approaches, but in multi-echo approaches, they are visible in the form of alternative echoes and thus provide valuable information.
Second, multi-echo provides information if multiple objects are located at the same observation angle, which can be useful for determining the noise caused by adverse weather.
With these insights, the promise of our approach is increased and more reliable LiDAR information in adverse weather conditions and thereby we increase the safety and performance of downstream tasks.

It is a nontrivial task to determine, which echoes are useful since most do not represent any physical object, but rather are artifacts from reflections or refractions.
In this work, we present a novel self-supervised approach \textbf{SMED}Net (\textbf{S}elf-supervised-\textbf{M}ulti-\textbf{E}cho-\textbf{D}enoiseNet) to achieve this goal.
More specifically, the task is to pick an echo for each laser pulse that is most likely representing an object of interest and discard other echoes.
The resulting point cloud has a single echo per discrete observation angle.
Although multiple echoes can represent an object of interest, the output is defined as a traditional single echo point cloud because many downstream tasks process point clouds in such a format.
Moreover, the single echo format is generally sufficient in clear weather conditions.
The explained task is illustrated in Fig. \ref{fig:idea}.
Finally, we propose a characteristics similarity regularization method to boost the performance of our self-supervised framework.
The proposed method improves both convergence and performance at test time. 

The contributions of this paper are summarized as follows:
\begin{itemize}
    \item We propose and formulate the task of multi-echo denoising for LiDAR in adverse weather.
    \item We propose a novel self-supervised approach SMEDNet that learns this task, motivated by the fact that point-wise labels are laborious to obtain especially for multi-echo point clouds.
    The approach is based on neighborhood correlation and a novel blind spot method, extending the work of Bae \textit{et al.} \cite{bae2022slide}.
    \item We propose the characteristics similarity regularization for LiDAR adverse weather denoising to boost the performance of our method.
\end{itemize}

\section{Related Work}
\label{sec:related_work}

\subsection{Classical methods}

Several classical methods have been presented in the literature for denoising adverse weather-corrupted point clouds.
They typically exploit the fact that the expected density of noise caused by adverse weather is low \textit{i.e.} the noisy points are more sparse compared to valid points.
Rönnbäck \textit{et al.} proposed an approach based on median filtering of the LiDAR range measurements corrupted by snowfall \cite{ronnback2008filtering}.
Median filtering showed a substantial reduction in snowflake detects.
Radius outlier removal (ROR) and statistical outlier removal (SOR), proposed by Rusu \textit{et al.} \cite{rusu20113d}, remove points based on their distance to neighboring points.
A constant radius threshold based on statistics defines whether a point is an inlier or an outlier.
However, these methods do not perform well with LiDAR data as it has drastically varying point density.
This leads to ROR and SOR filters being biased on the measured range. 
Charron \textit{et al.} \cite{charron2018noising} presented an algorithm called dynamic radius outlier removal (DROR).
It solves the problem of varying density via a dynamic radius threshold.
The threshold is adjusted based on the measured range and thus range induced bias is eliminated.
Park \textit{et al.} \cite{park2020fast} discovered that adverse weather, specifically snowfall, caused points to have typically lower intensity compared to valid points.
They developed low-intensity outlier removal (LIOR), which removes points noise points based on the intensity value and ROR output. 
Later on, dynamic distance–intensity outlier removal (DDIOR) presented by Wang \textit{et al.} \cite{wang2022scalable} fuses LIOR and DROR to improve the performance.
More recently, Li \textit{et al.} proposed to use spatiotemporal features to identify points caused by adverse weather \cite{li2022snowing}.
They utilized the phenomenon that adverse weather-induced points follow unpredictable trajectories. 

\subsection{Learned methods}


Deep learning has been established as a versatile framework, and it has been shown to be successful in a multitude of tasks, including adverse weather denoising.
Because adverse weather denoising can be thought of as a semantic segmentation task, general-purpose segmentation networks can be utilized.
Most popular general-purpose approaches use voxelized \cite{tang2020searching, zhu2021cylindrical}, bird's eye view \cite{zhou2021panoptic}, or spherical projection input \cite{wu2018squeezeseg, wu2019squeezesegv2, xu2020squeezesegv3, razani2021lite}.
The raw point cloud approaches exist as well \cite{qi2017pointnet, qi2017pointnet++, landrieu2018large, su2018splatnet, tatarchenko2018tangent, hu2020randla, rosu2022latticenet, yan2021sparse}.
While these general-purpose semantic segmentation approaches can be utilized to denoise point clouds corrupted by adverse weather, they are not designed for the given task, thus they can lack performance.

Parallels can be drawn between adverse weather denoising and deep image denoising where camera images are restored as these methods modify the original pixels and try to acquire a clean image.
This can be done supervised or self-supervised manner.
One of the pioneering works is the Noise2Noise \cite{lehtinen2018noise2noise}, which maps a noisy image to another noisy image with the same underlying signal. 
Later on, methods that do not require image pairs have been developed \cite{laine2019high, quan2020self2self, huang2021neighbor2neighbor, wang2022blind2unblind}.
Therefore, these methods tend to have more flexible use cases.
Despite the success of deep image denoising, the methods do not directly suit the purpose of LiDAR adverse weather denoising as the data is sparse and the task is to remove noise points and keep valid points unaltered.
Some works have developed dense point cloud denoising methods \cite{hermosilla2019total, luo2020differentiable}, which have an equivalent goal to deep image denoising.
However, these methods are yet to be implemented on the adverse weather denoising task.

Deep learning methods specifically targeted to the adverse weather denoising task have been also presented in the literature, and they are typically projection-based as it provides a good trade-off between accuracy and computational cost.
Heinzler \textit{et al.} \cite{heinzler2020cnn} proposed the WeatherNet architecture for denoising fog and rain.
Seppänen \textit{et al.} introduced 4DenoiseNet \cite{seppanen20224denoisenet}, which utilizes spatiotemporal information from consecutive point clouds and proposed a semi-synthetic dataset for training without human-annotated data.
Bae \textit{et al.} \cite{bae2022slide} developed SLiDE, which can be trained in a self-supervised manner via reconstruction difficulty.
Yu \textit{et al.} \cite{yu2022lisnownet} used the fast Fourier and the discrete wavelet transforms to construct a loss function for training a network for snow segmentation.
Our work is inspired by the network architecture of \cite{seppanen20224denoisenet} and the self-supervised pipeline of \cite{bae2022slide}, and builds the multi-echo denoising framework on them.

Preliminary work has been done on airborne particle classification using multi-echo point clouds.
Stanislas \textit{et al.} \cite{stanislas2021airborne} studied supervised deep learning methods for classifying airborne particles using the multi-echo measurement as a feature.
That is, they predicted the class only for the first echo and used the alternative echo merely as a feature. 
This approach has the limitation of not utilizing alternative echoes to their fuller potential, as they can be used for viable substitute points to replace points of the strongest echo point cloud, which is the goal and contribution of our work.
We are the first ones to present a method that picks substitutes from a multi-echo point cloud to replace invalid points of the strongest echo point cloud.

\section{Methods}
\label{sec:methods}

\noindent \textbf{Multi-echo input formulation}.
A traditional single-echo point cloud is defined as $\mathbf{P}_s \in \mathbb{R}^{N_p \times N_c}$, where $N_p$ and $N_c$ denote the number of points and channels, respectively.
A multi-echo point cloud is defined as $\mathbf{P}_m \in \mathbb{R}^{N_p \cdot N_e \times N_c}$, where $N_e$ denotes the number of echoes per emitted laser pulse.
In this work, point clouds are processed in an ordered format.
We define multi-echo ordered format as a spherical projection of the point cloud $\Gamma : \mathbb{R}^{N_p \cdot N_e \times N_c} \to \mathbb{R}^{H \times W \times N_e \times N_c}$, where $H$ and $W$ stand for image dimensions of the projection. 
As echoes from the same laser pulse have the same azimuth and elevation angles, they are stacked on dimension $N_e$, and a vector along this dimension is also denoted as an echo group $E_g$.
Finally, the multi-echo ordered point cloud is $\mathbf{P}_{meo} \in \mathbb{R}^{H \times W \times N_e \times N_c}$.

\noindent \textbf{Multi-echo denoising task formulation}.
The intuition behind multi-echo denoising is to see through the noise caused by \textit{e.g} adverse weather by recovering points that carry useful information from other echoes.
Additionally, misleading or irrelevant points caused by other echoes are to be discarded.
To formulate this task, let $M(\mathbf{P}_{nm}) = \mathbf{P}_{c}$, where $\mathbf{P}_{nm} \in \mathbb{R}^{N_p \cdot N_e \times N_c}$ is the noisy multi-echo point cloud and $\mathbf{P}_c \in \mathbb{R}^{(N_p \cdot N_e - N_n - N_r) \times N_c}$ is the obtained clean point cloud. 
$N_n$ denotes the number of removed noise points, and $N_r$ denotes other irrelevant points, for instance, duplicate, and artifact points caused by the refraction of the laser.
The remaining points $\mathbf{P}_c$ include substitutes recovered from alternative echoes.
Ultimately the goal is to obtain $\mathbf{P}_c$ such that it is equivalent to a standard strongest echo point cloud in clear weather, as it is the default singe-echo mode.
In this work, $M(\cdot)$ is defined as a self-supervised neural network, which is described in the following subsections.

\subsection{SMEDNet}

\begin{figure}
\centering
\includegraphics[width=0.48\textwidth]{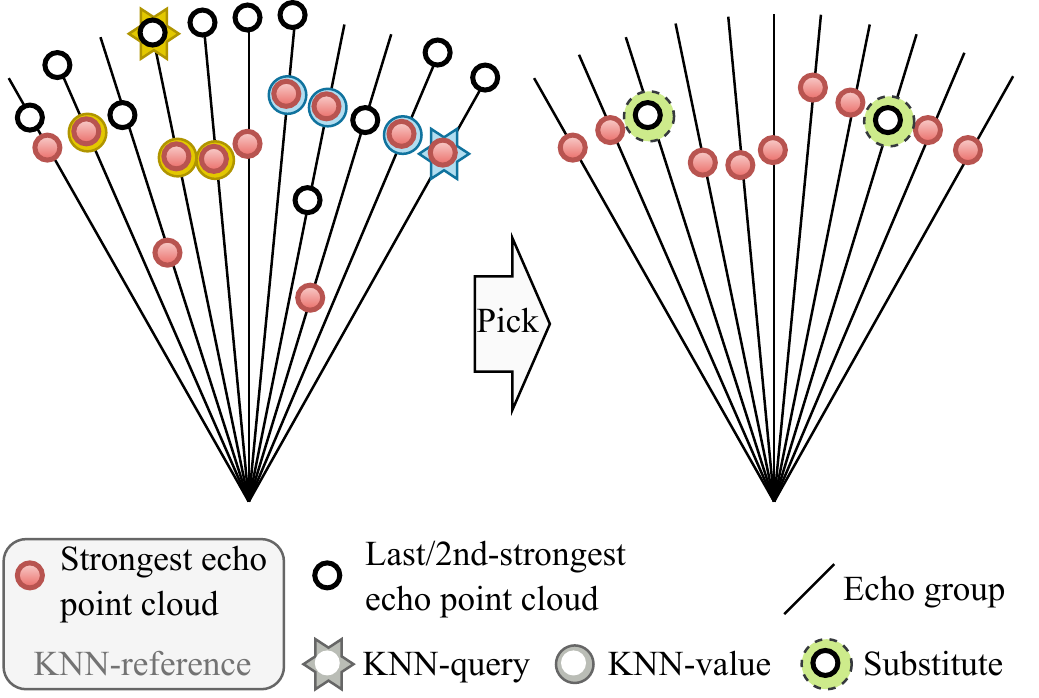}
\caption{In the proposed multi-echo neighbor encoder, KNN sets are computed using multi-echo query and KNN reference, we use the strongest echo point cloud as it is \textit{de facto} in single-echo approaches. 
Two example queries illustrate how KNN sets look. 
"Exclude self -- Include neighbors" simply discards KNN queries and keeps the values. 
From this, the coordinate learner predicts the queries. 
The point cloud on the right illustrates the final output of our method, where substitutes are points from alternative echoes.}
\label{fig:knn-multi-echo}
\end{figure}

\noindent \textbf{Multi-echo neighbor encoder.} 
The multi-echo ordered point cloud $\mathbf{P}_{meo}$ is the input to the multi-echo neighbor encoder, which is one of our contributions and the main difference to the work of Bae \textit{et al.} \cite{bae2022slide}. 
This module processes $\mathbf{P}_{meo}$ into upper bound KNN sets using Euclidean distance to the reference point cloud \textit{i.e.} strongest echo point cloud and encodes these values into a feature tensor $\mathbf{\Pi}$.
The reference point cloud is the strongest point cloud given that it is the default in clear weather, and thus most likely to contain points of objects of interest.
This is illustrated in Fig. \ref{fig:knn-multi-echo}.
We formulate the multi-echo neighbor encoder as a set of convolutions.
For simplicity, components of $\mathbf{P}_{meo}$ are denoted as: $\mathbf{P}_{xyz}$ -- Cartesian coordinates, $\mathbf{P}_{\theta \phi}$ -- azimuth and elevation coordinates, and $\mathbf{P}_{r}$ -- range coordinates.
\begin{multline}
    \mathbf{\Pi} = \mathbf{w}_t * \mathbf{P}_{meo} \\
    = \mathbf{w}_t (\mathbf{P}_{r}[\mathit{argmink}_{\Vec{p}_2}(\psi(\Vec{p}_2, \Vec{p}_3))] \\
    \oplus (\mathbf{P}_{\theta \phi}[\Vec{p}_2] - \mathbf{P}_{\theta \phi}[\mathit{argmink}_{\Vec{p}_2}(\psi(\Vec{p}_2, \Vec{p}_3))]))
\end{multline}
\noindent where $\mathbf{w}_t \in \mathbb{R}^{k \cdot N_e \times S_o}$ indicates trainable weights, $\Vec{p}_2 = (h, w, 0)$, $h \in [\![0, H]\!]$, and $w \in [\![0, W]\!]$ indicates a pixel coordinate on the strongest echo point cloud.
$argmink_{\Vec{p}_2}(\cdot)$ returns indices of $k \times N_e$ strongest echo values which minimize Euclidian distance to multi-echo queries $\Vec{p}_3 = (h, w, \hat{e}), \hat{e} \in [\![0, N_e]\!]$.
$\oplus$ is the concatenation operation, and 
\begin{multline}
    \psi(\Vec{p}_2, \Vec{p}_3) = ||\mathbf{P}_{xyz}[\Vec{p}_2 + \Delta \Vec{p}_2] - \mathbf{P}_{xyz}[\Vec{p}_3]||_2 \\
    \Delta \Vec{p}_2 \in \mathbf{A}_c \qquad \forall \psi(\Vec{p}_2, \Vec{p}_3) < C_r 
\end{multline}
\noindent where $C_r$ is a fixed radius cutoff hyper-parameter defining the upper bound for the neighbor search, which ensures that only local points are considered. 
$\mathbf{A}_c$ defines the elements considered in the search.
$\mathbf{P}_{\theta \phi}$ are encoded to preserve the 3D information because the grid positions of the neighbors $\mathbf{P}_{meo}$ are lost due to the nature of the KNN search.
Hereby, the architecture can utilize the original 3D information for the predictions.
The final output of the module is the activated features. 
Next, the coordinate and correlation learner models process these features.

\noindent \textbf{Coordinate learner.} 
The task of the coordinate learner is to predict the coordinate of a point $p_i$ given the neighboring points of $p_i$.
To simplify the task and enable faster convergence, the three dimensions $p_i \in \mathbb{R}^{3}$ are reduced to one $r_i = ||p_i||_2 \in \mathbb{R}$.
The coordinate learner learns point coordinates by minimizing the absolute distance error to the actual coordinates.

\noindent \textbf{Correlation learner.} 
The task of the correlation learner is to predict the predictability of the coordinate of point $p_i$.
More precisely, the task is to predict inversely the correlation to the neighboring points, hence the name correlation learner.
This is done because the loss is formulated in such a manner that the correlation learner predicts a high value for $p_i$ if its coordinate is difficult to predict \textit{i.e.} to compensate for a high coordinate prediction error.
During training this network learns to identify points that are caused by objects of interest, assuming that those points have a higher neighborhood correlation compared to other points.
During inference time the points can be filtered based on the predicted value of the correlation learner, thus only the correlation learner is used during inference time (Section \ref{sec:inference}).

\noindent \textbf{Exclude self -- Include neighbors.} 
The key difference between the coordinate and correlation learner is that blind spots are added to the input of the coordinate learner during training.
Here we present a novel variant of the blind spot method.
Previously, it has been implemented on deep image denoising \cite{laine2019high, krull2019noise2void, wang2022blind2unblind}, where the center pixel of a traditional convolution filter is blanked.
Here we combine the multi-echo neighbor encoder with the blind-spot approach.
That is, for a $p_i$ the input is its KNN set without the query (Fig. \ref{fig:knn-multi-echo}).
To learn the coordinate of $p_i$, only its neighbors are processed and the coordinate of $p_i$ is predicted.
However, to learn the correlation of $p_i$, the whole KNN set including the query, is processed.
This ensures that the model can utilize the valuable information of the query point.
The benefit of our method is that physically neighboring data points are more relevant when estimating the value of the hidden point compared to prior work where more unrelated 2D grid relations are used.
Following \cite{quan2020self2self}, we pick a random subset of points to be excluded in order to avoid over-fitting. 

\subsection{Characteristics similarity regularization}


To boost the performance of our method, we present a novel characteristics similarity regularization (CSR) method for the adverse weather denoising task.
This method accelerates the convergence of the self-supervised learning regime and increases the performance of the learned model.
Notably, it requires only \textit{one} hyper-parameter, the size of the search $k_{CSR}$.
The effect of $k_{CSR}$ on performance is small, thus it is easy to tune.
The CSR is built on the following assumptions of the nature of the noise caused by adverse weather.
1) An expected echo caused by an airborne particle has a typical intensity \cite{o1970visibility}.
2) Airborne particles cause more sparse point clouds compared to other objects \cite{charron2018noising}.
Based on these assumptions, a process that guides the predictions into similar distribution as the intensity and sparsity should increase the convergence.
With this insight in mind, we build a regularization process to guide the predictions to the distribution mentioned above. 
\footnote{It is important to note here that we do not assume any specific distribution of the characteristics of the noise points, but instead encourage the model to learn the connections between these characteristics and other features relative to the output.}

Let us define the components of the characteristics map $\mathbf{\Theta}$.
The map has the same spatial dimensions as the output of the network, which enable straightforward implementation of the CSR.
As for point-source radiation, the expected intensity measurement from a LiDAR follows the inverse-square law $i \propto \frac{1}{r^2}$, where $r$ is the distance of the measurement.
To eliminate the contribution of range, the intensity matrix is defined as

\begin{equation}
  \mathbf{I} = \mathbf{I}_{raw} \odot \mathbf{P}_{r}^2
  \label{eq:intesity}
\end{equation}

\noindent where $\mathbf{P}_r = ||\mathbf{P}_{xyz}||_2$.
The sparsity is defined as follows,

\begin{equation}
  \mathbf{S} = E_d(\mathbf{P}_{meo}) \odot \frac{1}{\mathbf{P}_r}
  \label{eq:sparsity}
\end{equation}

\noindent where $E_d(\cdot)$ returns Euclidean distance to the nearest neighbor for each point.
Thus, sparsity can be obtained as a \textit{free lunch} from $\mathbf{\Pi}$.
It is normalized with the range matrix $\mathbf{P}_r$, as sparsity $s \propto r$.
Then, we get the characteristics map as a concatenation of the components $\mathbf{\Theta} = \mathbf{I} \oplus \mathbf{S}$.
The similarity is computed using normal distributions of neighbors of $\mathbf{\Theta}$.
That is, for each element in $\mathbf{\Theta}$, arguments of $k_{CSR}$-nearest-neighbors are computed using the Euclidean distance.
A regression goal for a corresponding output is the Z-score of the distribution collected with the neighbor search.
This regression goal forms the characteristics similarity loss term using the absolute error.
The described regularization process forces those correlation learner outputs $\mathbf{O}_{cor}$ that have similar characteristics to have similar values, which we expect will encourage the model to converge.
Finally, we summarize CSR as the Algorithm (\ref{alg:csr}).

\begin{algorithm}
\caption{Characteristics similarity regularization} \label{alg:csr}
\begin{algorithmic}
\For{$\text{a batch} \in \text{batches}$}
\State $\mathbf{I} \gets \text{Equation (\ref{eq:intesity})}$ \Comment{Intensity}
\State $\mathbf{S} \gets \text{Equation (\ref{eq:sparsity})}$ \Comment{Sparsity}
\State $\mathbf{\Theta} \gets \textproc{concat}(\mathbf{I}, \mathbf{S})$ \Comment{Characteristics map}
\State $\mathbf{\Phi} \gets \mathbf{O}_{cor}[\textproc{argNeigbors}(\mathbf{\Theta}, k_{CSR})]$
\State $\mathbf{\Xi} \gets |\textproc{Zscore}(\mathbf{O}_{cor}, \mathbf{\Phi})|$ \Comment{Characteristics \\ \hfill similarity loss}
\EndFor
\end{algorithmic}
\end{algorithm}

\subsection{Architecture and loss function}

\begin{figure*}
\centering
\includegraphics[width=1.0\textwidth]{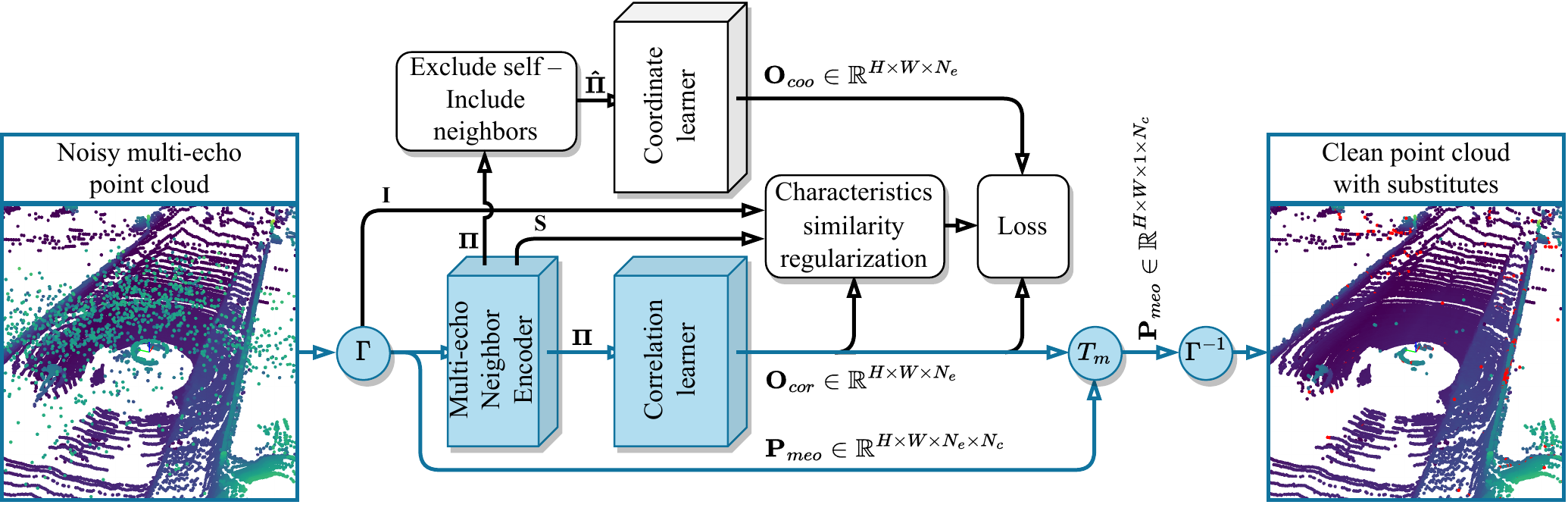}
\caption{The proposed self-supervised multi-echo denoising architecture (SMEDNet). 
White modules are used for training only.}
\label{fig:pipeline}
\end{figure*}

\noindent \textbf{Architecture.} 
Our proposed architecture is presented in Fig. \ref{fig:pipeline}.
The architecture consists of two encoder-decoder neural networks.
The networks are identical except for the input since the coordinate learner processes point clouds with blind spots.
The encoder-decoder is to close resemblance to 4DenoiseNet \cite{seppanen20224denoisenet} using three residual blocks.
Please refer to the supplementary paper for more details.

\noindent \textbf{Loss function.} 
Similarly to \cite{bae2022slide}, two networks are trained jointly. 
As stated above, the "exclude self -- include neighbors" is performed only for a randomly selected subset of points, denoted by $\mathbf{P}_s$.
Therefore, we train only those points.
The loss function in its final form is defined as follows:

\begin{equation}
  \mathcal{L} = \frac{1}{|\mathbf{P}_s|} \sum _{p_s \in \mathbf{P}_s} \left( \frac{\lambda \cdot |\mathbf{O}^{coo}_{p_s} - \mathbf{P}^{r}_{p_s}|}{\lceil \mathbf{P}^{r} \rceil _{p_s} \odot \textproc{exp}(\mathbf{O}^{cor}_{p_s})} + \mathbf{O}^{cor}_{p_s} + \mathbf{\Xi}_{p_s} \right)  
  \label{eq:loss}
\end{equation}

\noindent where $\mathbf{O}^{coo}$ and $\mathbf{O}^{cor}$ are the outputs of the coordinate and correlation learner, respectively.
The loss function minimizes the coordinate prediction error at the numerator and the denominator forces the output $\mathbf{O}^{cor}$ to a high value when the coordinate is difficult to predict \textit{i.e.} noisy adverse weather points.
Thus, during inference, $\mathbf{O}^{cor}$ is used as a score to define the noisy points. 
A fixed hyper-parameter $\lambda$ scales this term relative to the other terms.
A regulating term $+ \mathbf{O}^{cor}$ prevents predictions from exploding. 
The division by $\lceil \mathbf{P}^r \rceil$ reduces the range-induced bias of the learning process as the correlation to neighbors is inversely proportional to the range.
To make the learning process more stable, $\lceil \mathbf{P}^r \rceil$ is rounded up to full meters.
The model converges when the coordinate learner has learned meaningful high-level features of valid points and the correlation learner learns to output a high score, \textit{i.e.} low correlation value, for invalid points.

\subsection{Inference mode} \label{sec:inference}

During the inference mode, only the correlation learner is used.
The output $\mathbf{O}_{cor}$ is processed into class labels.
For this purpose, we formulate the multi-echo denoising classes in the following manner:

\begin{itemize}
    \item valid strongest echo, $(S \land T) \to VS$,
    \item potential substitute, $(\lnot S \land T \land B \land D) \to PS \in E_g$,
    \item discarded, $(\lnot VS \lor \lnot PS) \to DI$,
\end{itemize}

where $S, T, B, D$ are boolean formulas for "the strongest echo", "satisfies a threshold", "the best score", and "a different coordinate to strongest", respectively. 
$E_g$ denotes the echo group \textit{i.e.} the values along the $N_e$-dimension of $\mathbf{O}_{cor}$.
Here, we formulate $B$ in general form, but in our case, it is equal to the predicted correlation value $\mathbf{O}_{cor}$.
During the inference time, the $DI$ labeled points are removed ($T_m$ in Fig. \ref{fig:pipeline}), and the remaining point cloud is the final output of our method.

\section{Experimental results}
\label{sec:experiments}

\subsection{Implementation details} \label{sec:implementation}

\noindent \textbf{Datasets.}
The experiments are conducted with the STF dataset \cite{Bijelic_2020_STF}, which includes multi-echo LiDAR point clouds from snowfall, rain, and fog.
This is a suitable dataset for our experiments because it has a wide variance of adverse weather conditions in traffic scenarios which are our main interest.
STF \cite{Bijelic_2020_STF} is used only for qualitative experiments as it does not have point-wise labels.
Therefore, a semi-synthetic single-echo SnowyKITTI dataset \cite{seppanen20224denoisenet} is used for the quantitative experiments, which is a well-suited benchmark for comparison and ablation studies.

\noindent \textbf{2.5-echo point cloud.}
The LiDAR used in this study provides the strongest and last echo point clouds.
If the last echo is the same as the strongest echo, it is replaced by the second strongest echo.
Thus, we call it a 2.5-echo point cloud. 
Hypothetically, more than 2.5 echoes are beneficial, as those could represent objects of interest. 
Therefore, our method is formulated without a restriction on the number of echoes \textit{i.e.} the echo number has a dedicated tensor dimension and the operations are agnostic to the size of this dimension.
One case where the last echoes are useful is when the strongest echo is caused by adverse weather and the last echo is caused by the object of interest in the same echo group. 

\noindent \textbf{Training and testing details.}
The hyper-parameter settings were selected with a grid search as follows.
The learning rate is 0.01, the learning rate decay is 0.99 to avoid over-fitting, the fixed hyper-parameter in Eq. (\ref{eq:loss}) $\lambda$ = 5, and the $k_{CSR} = 9$.
We use the stochastic gradient descent optimizer with a momentum of 0.9 and train for 30 epochs.
The train/validate/test-set split ratio is 55/10/35, where all sets have a wide range of adverse weather corrupted point clouds but are from different sequences. 
During inference, a threshold $T_n = 0$ is empirically found to yield the best results.
The models are run on an RTX 3090 GPU using Python 3.8.10 and PyTorch 1.12.1 \cite{paszke2019pytorch}.

\subsection{Quantitative results}
\label{sec:n_res}

Quantitative experiments are conducted with the semi-synthetic SnowyKITTI \cite{seppanen20224denoisenet} LiDAR adverse weather dataset as labeled multi-echo data is not available.
To enable our algorithm to work on single-echo data, only the strongest echo is used.
Single-echo performance on the SnowyKITTI-dataset \cite{seppanen20224denoisenet} can be seen in Table \ref{performance}.
Our algorithm is compared to the state-of-the-art self-supervised SLiDE \cite{bae2022slide} and with well-proven classical methods DROR \cite{charron2018noising} and LIOR \cite{park2020fast}.
The methods are evaluated with the widely adopted Intersection over Union (IoU) metric.
The results are presented separately for the light, medium, and heavy levels of snowfall to have a more detailed indication of performance. 
Based on the results, the performance varies depending on the noise level.
Our method achieves superior performance to these methods in all levels of noise.

We also report the runtime and parameter count as adverse weather denoising methods are typically intended for a pre-processing step before other tasks, thus they add latency and memory usage to the pipeline.
Therefore, attention should be also given to these metrics. 
As indicated by the results, our method has approximately the same runtime and uses 35\% fewer parameters compared to SLiDE \cite{bae2022slide}.
This is significant as fewer parameters help with over-fitting, reduce memory footprint, and enable faster convergence.
A slight difference in the runtime between our method and SLiDE \cite{bae2022slide} results from the input layer, as the neighbor encoder is slower than a standard 2D convolution.
Overall, our method achieves superior performance and thus sets the new state-of-the-art in self-supervised adverse weather denoising.

\begin{table*}[!t]
\centering
\caption{Results on the SnowyKITTI \cite{seppanen20224denoisenet} dataset. * -- no training required. The bold font indicates the best values.}
\label{performance}
\begin{tabular}{ c c | c c c | c c }
 \toprule
 \multirow{2}{*}{Type} & \multirow{2}{*}{Method} & \multicolumn{3}{c}{IoU} & \multicolumn{1}{|c}{Runtime} & \multicolumn{1}{c}{Param.} \\
 & & Light & Medium & Heavy & ms & $\cdot 10^6$ \\ 
 \midrule
 \midrule
 \multirow{2}{*}{Classical} & DROR* \cite{charron2018noising}
 & 0.451 & 0.450 & 0.439 & 120  & $10^{-5}$\\
 & LIOR* \cite{park2020fast}
 & 0.448 & 0.447 & 0.434 & 120  & $10^{-5}$\\
 \midrule
 \multirow{2}{*}{Self-supervised} 
 & SLiDE \cite{bae2022slide}
 & 0.775 & 0.622 & 0.741 & \textbf{2.0} & 1.73  \\ 
 & SMEDNet (Ours)
 & \textbf{0.933} & \textbf{0.854} & \textbf{0.843} & 2.2   & 1.13  \\ 
 \midrule
 & \textit{Improvement}
 & 0.158 & 0.232 & 0.102 & & \\
\bottomrule
\end{tabular}
\end{table*}

\subsection{Ablation study}
\label{sec:ablations}

To assess the contributions of the characteristics similarity regularization, cutoff radius $C_r$, the style of the input convolution, and the neural network, an ablation study was conducted.
Table \ref{ablation_table} presents the IoU measurements when ablating the aforementioned modules.
The ablation of CSR indicates that performance on the test set increases with the inclusion of the module.
Excluding $C_r$ has a significant effect on the performance as well.
We suspect that this is because limiting the input neighbors with a certain threshold excludes points that are irrelevant for the coordinate prediction.
KNN search is replaced by simply including the grid neighbors, which is equal to a standard 2D convolution.
The performance decrease is due to the fact that grid neighbors can be spatially unrelated to the center point.
The neural network was changed to WeatherNet \cite{heinzler2020cnn}, as 4DenoiseNet \cite{seppanen20224denoisenet} has shown better performance in supervised learning \cite{seppanen20224denoisenet}.
The network affects also self-supervised performance based on our experiment.

\begin{table}[!t]
\centering
\caption{Ablations of different modules. CSR is the characteristics similarity regularization, $C_r$ is the cutoff radius of the multi-echo neighbor encoder, and the NS indicates the style of the neighbor search, where KNN is k-nearest-neighbors and GridN is the grid neighbors. NN denotes a neural network, where 4DN is 4DenoiseNet \cite{seppanen20224denoisenet}, and WN is WeatherNet \cite{heinzler2020cnn}.}
\label{ablation_table}
\begin{tabular}{ c c c c | c c c }
 \toprule
 \multirow{2}{*}{CSR} & \multirow{2}{*}{$C_r$} & \multirow{2}{*}{NS} & \multirow{2}{*}{NN} & \multicolumn{3}{c}{IoU} \\
 & & & & Light & Medium & Heavy \\ 
 \midrule
 \midrule
 \xmark & \cmark & KNN    & 4DN    & 0.912 & 0.825 & 0.802 \\ 
 \cmark & \xmark & KNN    & 4DN    & 0.871 & 0.819 & 0.808 \\
 \cmark & \cmark & GridN  & 4DN    & 0.854 & 0.795 & 0.806 \\
 \cmark & \cmark & KNN    & WN     & 0.885 & 0.819 & 0.792 \\
 \midrule
 \cmark & \cmark & KNN    & 4DN    & 0.933 & 0.854 & 0.843 \\
\bottomrule
\end{tabular}
\end{table}

\subsection{Qualitative results and discussion}
\label{sec:q_res_and_d}

\begin{figure*}
     \centering
     \adjustbox{minipage=0.6em,valign=t}{\subcaption{}\label{fig:sample1}}%
     \begin{subfigure}{\dimexpr1\linewidth-1.3em\relax}
         \centering
         \begin{subfigure}[b]{0.19\textwidth}
            \caption*{Input}
            \frame{\includegraphics[width=0.98\textwidth]{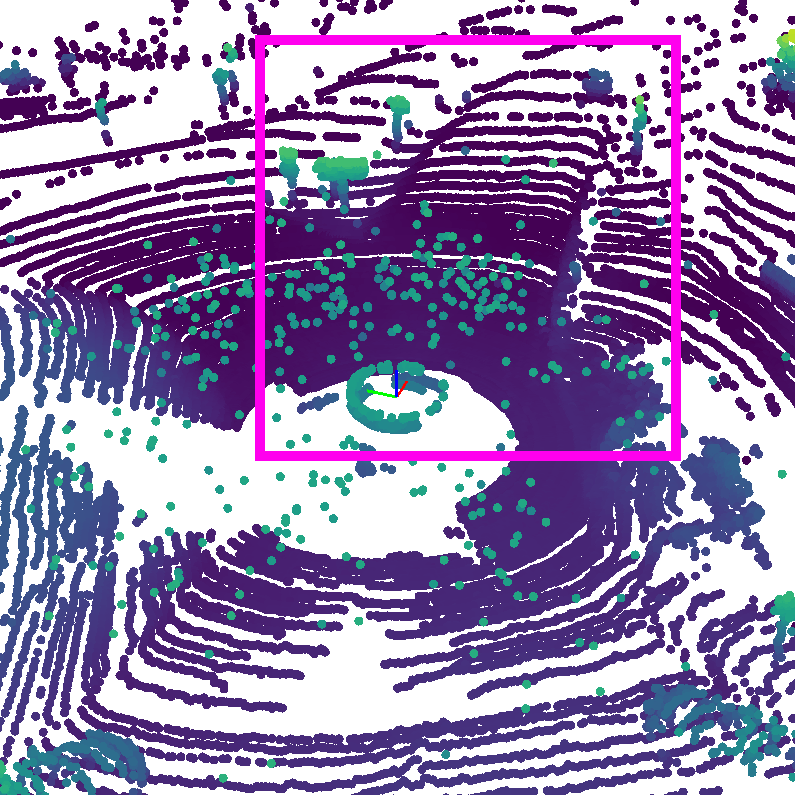}}
         \end{subfigure}
         \begin{subfigure}[b]{0.38\textwidth}
            \caption*{Baseline MEDROR}
            \frame{\includegraphics[width=0.49\textwidth]{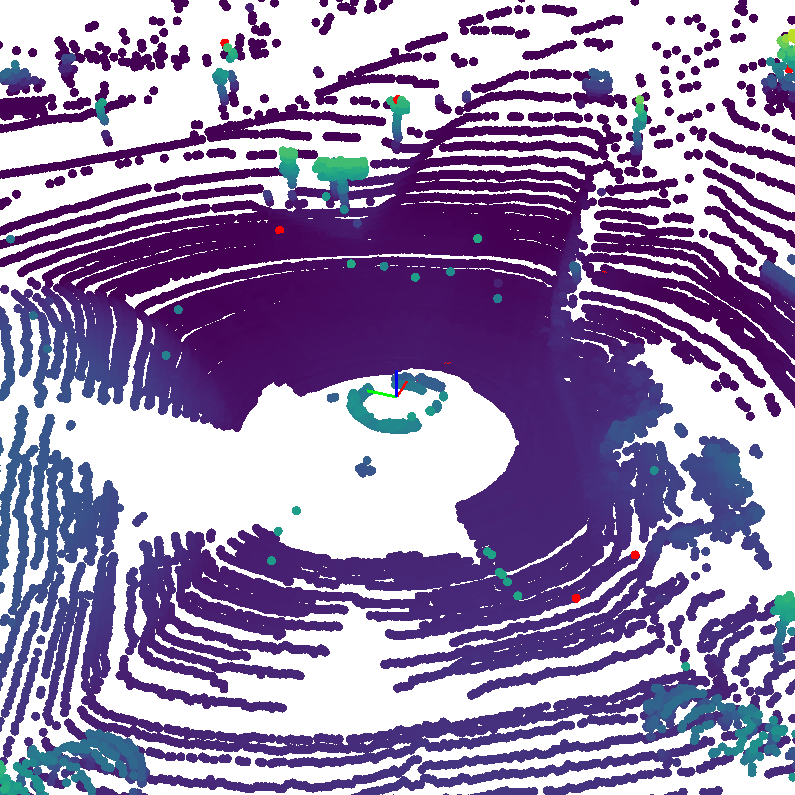}}
            \hspace{-1.5mm}
            \frame{\includegraphics[trim={260px 339px 119px 40px}, clip, width=0.46\textwidth, cfbox=Magenta2 1.5pt 1.5pt]{medror_seq4_174.png}}
         \end{subfigure}
         \begin{subfigure}[b]{0.38\textwidth}
            \caption*{Our SMEDNet}
            \frame{\includegraphics[width=0.49\textwidth]{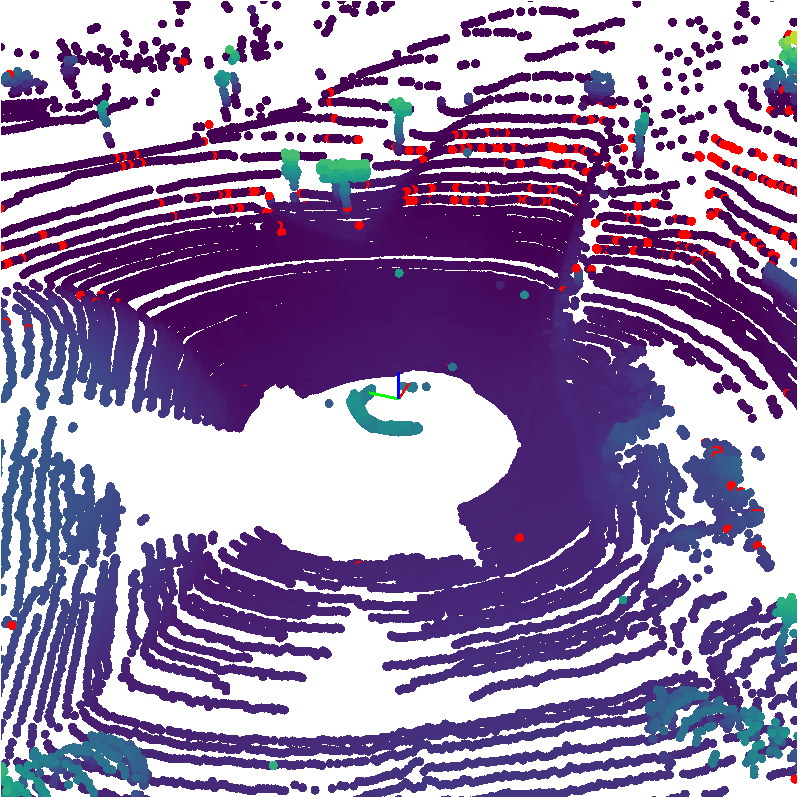}}
            \hspace{-1.5mm}
            \frame{\includegraphics[trim={260px 339px 119px 40px}, clip, width=0.46\textwidth, cfbox=Magenta2 1.5pt 1.5pt]{smednet_seq4_174.png}}
         \end{subfigure}
         \vspace{3.0mm}
     \end{subfigure}
     
     \adjustbox{minipage=0.6em,valign=t}{\subcaption{}\label{fig:sample2}}%
     \begin{subfigure}{\dimexpr1\linewidth-1.3em\relax}
         \centering
         \begin{subfigure}[b]{0.19\textwidth}
            \frame{\includegraphics[width=0.98\textwidth]{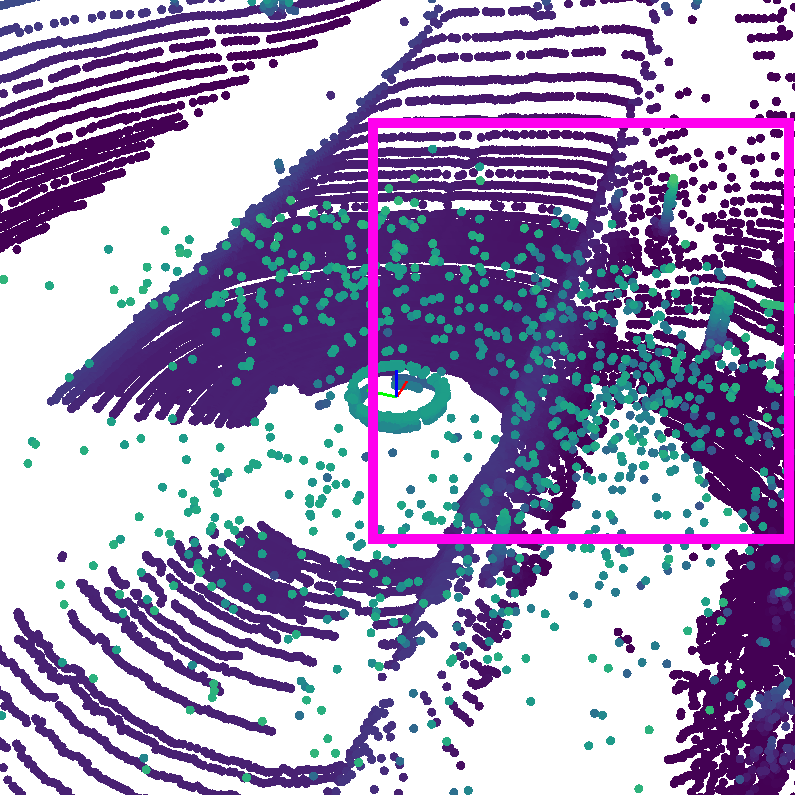}}
         \end{subfigure}
         \begin{subfigure}[b]{0.38\textwidth}
            \frame{\includegraphics[width=0.49\textwidth]{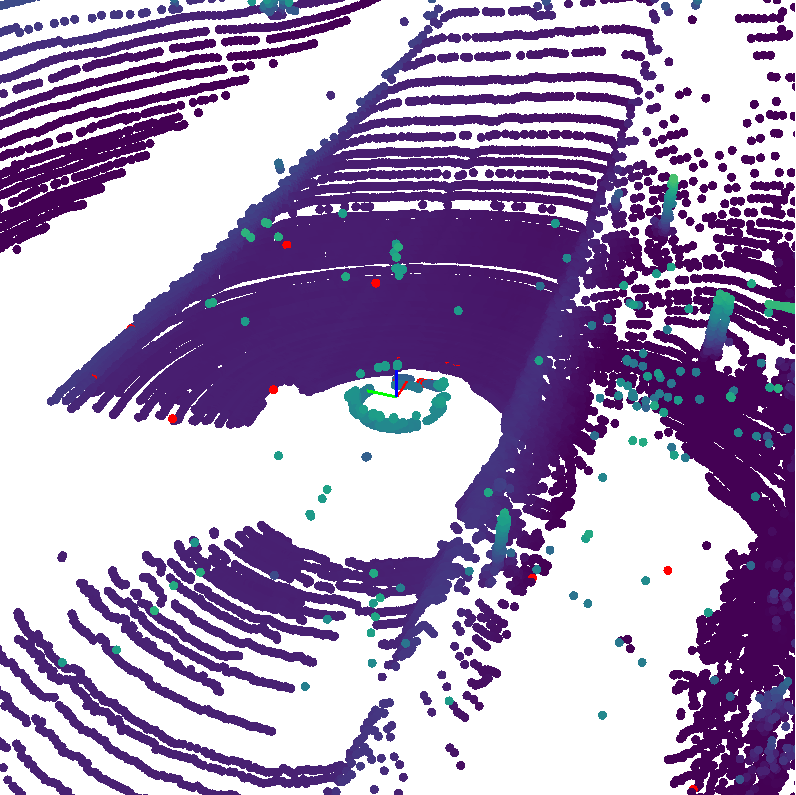}}
            \hspace{-1.5mm}
            \frame{\includegraphics[trim={373px 256px 7px 123px}, clip, width=0.46\textwidth, cfbox=Magenta2 1.5pt 1.5pt]{medror_seq4_194.png}}
         \end{subfigure}
         \begin{subfigure}[b]{0.38\textwidth}
            \frame{\includegraphics[width=0.49\textwidth]{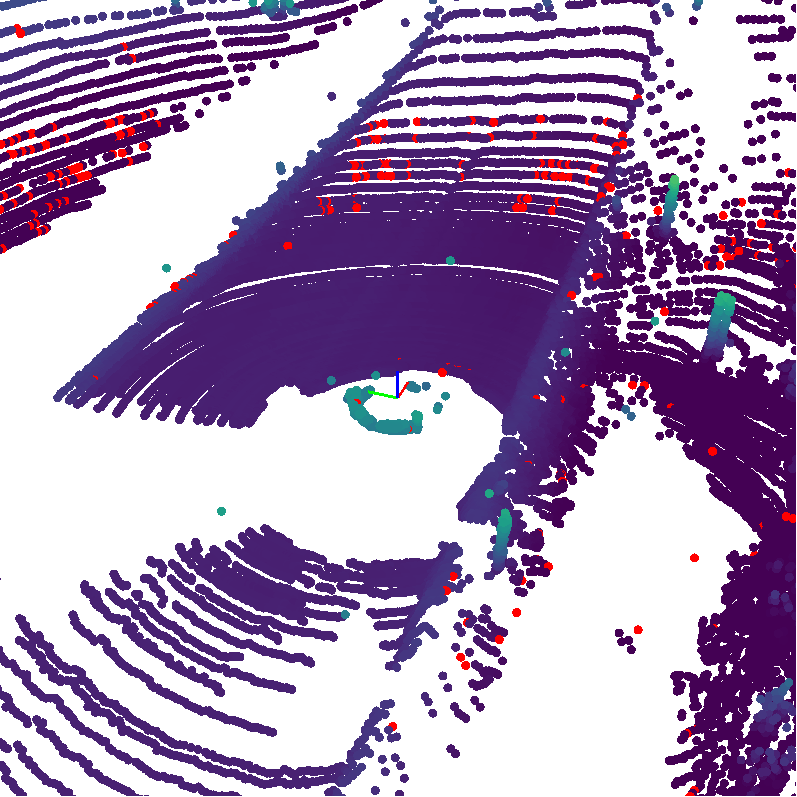}}
            \hspace{-1.5mm}
            \frame{\includegraphics[trim={373px 256px 7px 123px}, clip, width=0.46\textwidth, cfbox=Magenta2 1.5pt 1.5pt]{smednet_seq4_194.png}}
         \end{subfigure}
         \vspace{3.0mm}
     \end{subfigure}

     \adjustbox{minipage=0.6em,valign=t}{\subcaption{}\label{fig:sample3}}%
     \begin{subfigure}{\dimexpr1\linewidth-1.3em\relax}
         \centering
         \begin{subfigure}[b]{0.19\textwidth}
            \frame{\includegraphics[width=0.98\textwidth]{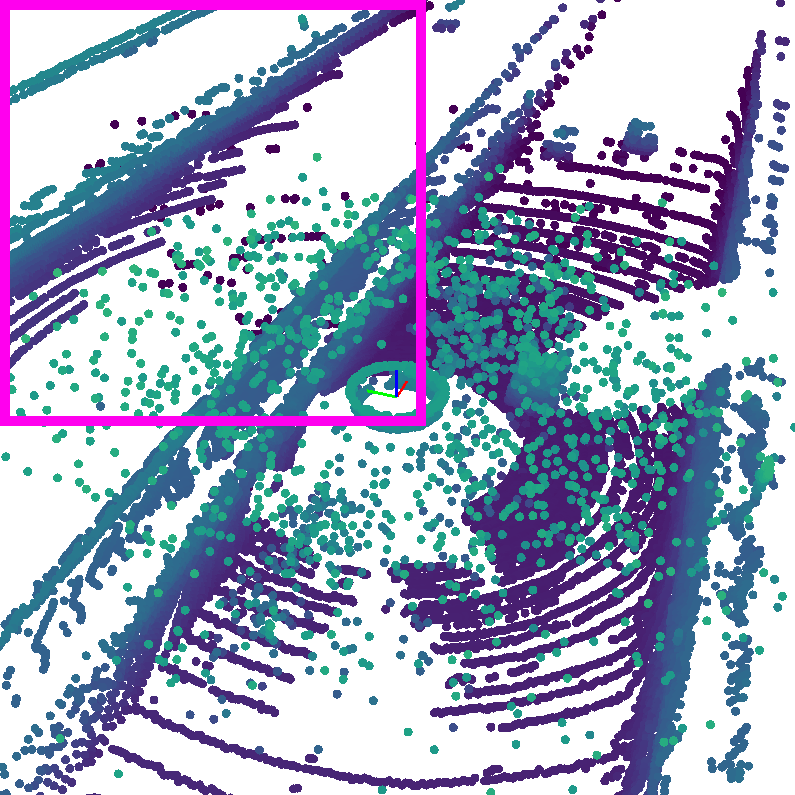}}
         \end{subfigure}
         \begin{subfigure}[b]{0.38\textwidth}
            \frame{\includegraphics[width=0.49\textwidth]{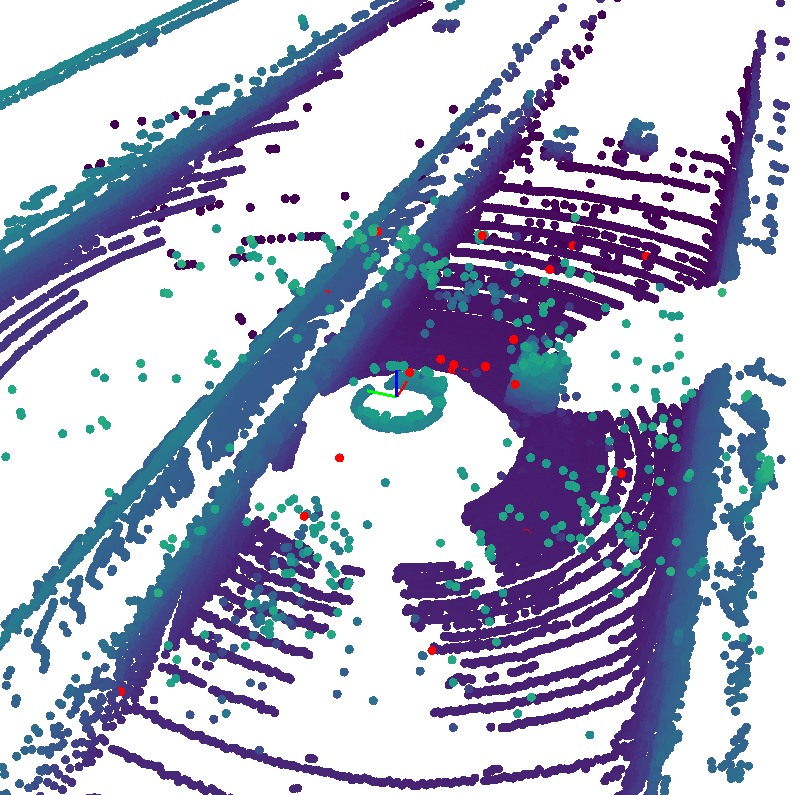}}
            \hspace{-1.5mm}
            \frame{\includegraphics[trim={0 379px 379px 0}, clip, width=0.46\textwidth, cfbox=Magenta2 1.5pt 1.5pt]{medror_seq4_421.png}}
         \end{subfigure}
         \begin{subfigure}[b]{0.38\textwidth}
            \frame{\includegraphics[width=0.49\textwidth]{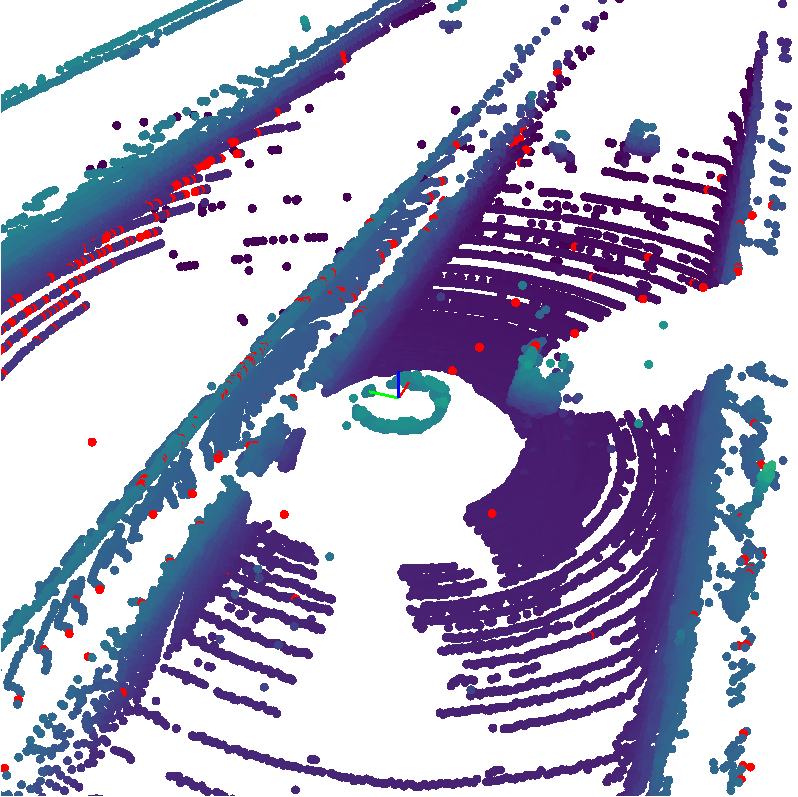}}
            \hspace{-1.5mm}
            \frame{\includegraphics[trim={0 379px 379px 0}, clip, width=0.46\textwidth, cfbox=Magenta2 1.5pt 1.5pt]{smednet_seq4_421.png}}
         \end{subfigure}
         \vspace{3.0mm}
     \end{subfigure}
     
     \adjustbox{minipage=0.6em,valign=t}{\subcaption{}\label{fig:sample4}}%
     \begin{subfigure}{\dimexpr1\linewidth-1.3em\relax}
         \centering
         \begin{subfigure}[b]{0.19\textwidth}    
            \frame{\includegraphics[width=0.98\textwidth]{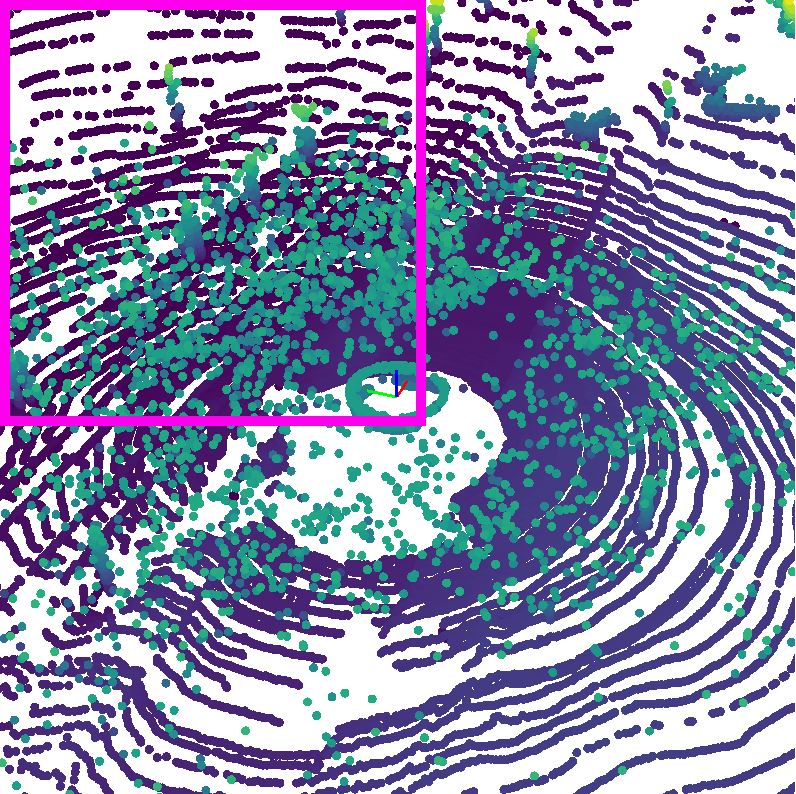}}
         \end{subfigure}
         \begin{subfigure}[b]{0.38\textwidth}
            \frame{\includegraphics[width=0.49\textwidth]{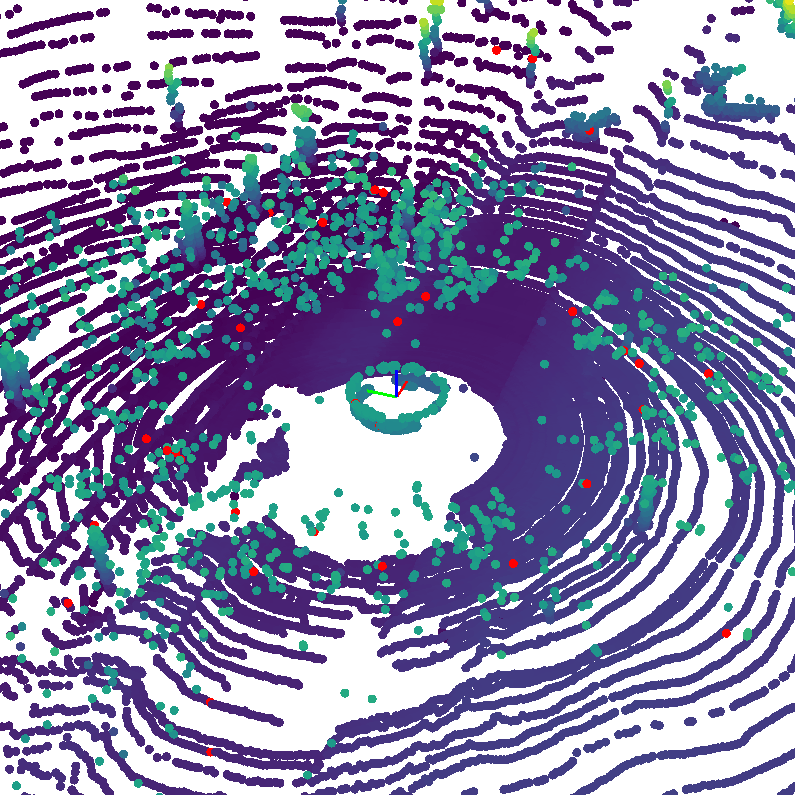}}
            \hspace{-1.5mm}
            \frame{\includegraphics[trim={0 379px 379px 0}, clip, width=0.46\textwidth, cfbox=Magenta2 1.5pt 1.5pt]{medror_seq4_364.png}}
         \end{subfigure}
         \begin{subfigure}[b]{0.38\textwidth}
            \frame{\includegraphics[width=0.49\textwidth]{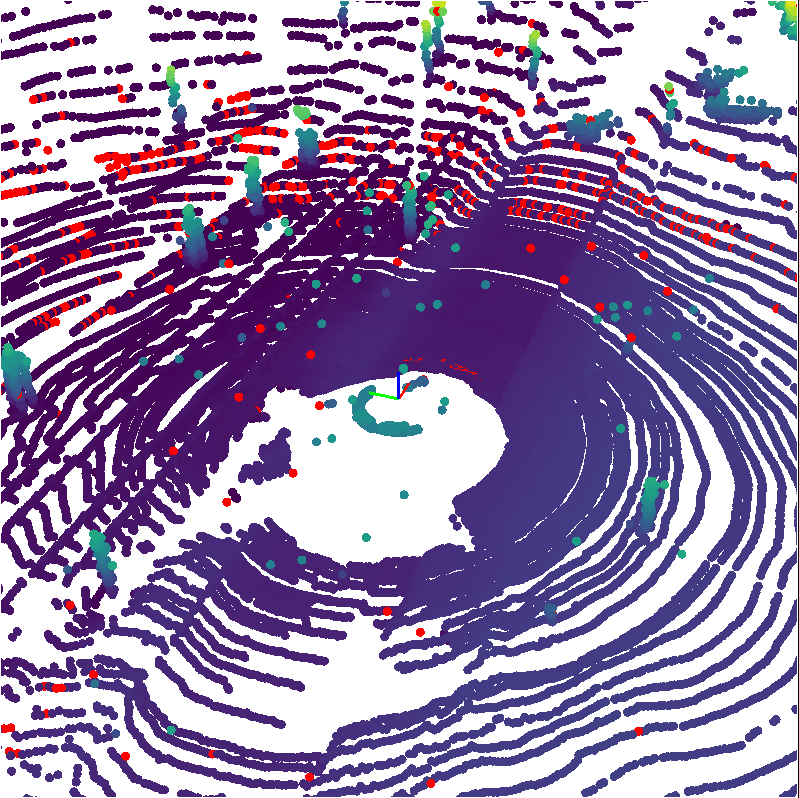}}
            \hspace{-1.5mm}
            \frame{\includegraphics[trim={0 379px 379px 0}, clip, width=0.46\textwidth, cfbox=Magenta2 1.5pt 1.5pt]{smednet_seq4_364.png}}
         \end{subfigure}
         \vspace{3.0mm}
     \end{subfigure}
     
     \adjustbox{minipage=0.6em,valign=t}{\subcaption{}\label{fig:sample5}}%
     \begin{subfigure}{\dimexpr1\linewidth-1.3em\relax}
         \centering
         \begin{subfigure}[b]{0.19\textwidth}
            \frame{\includegraphics[width=0.98\textwidth]{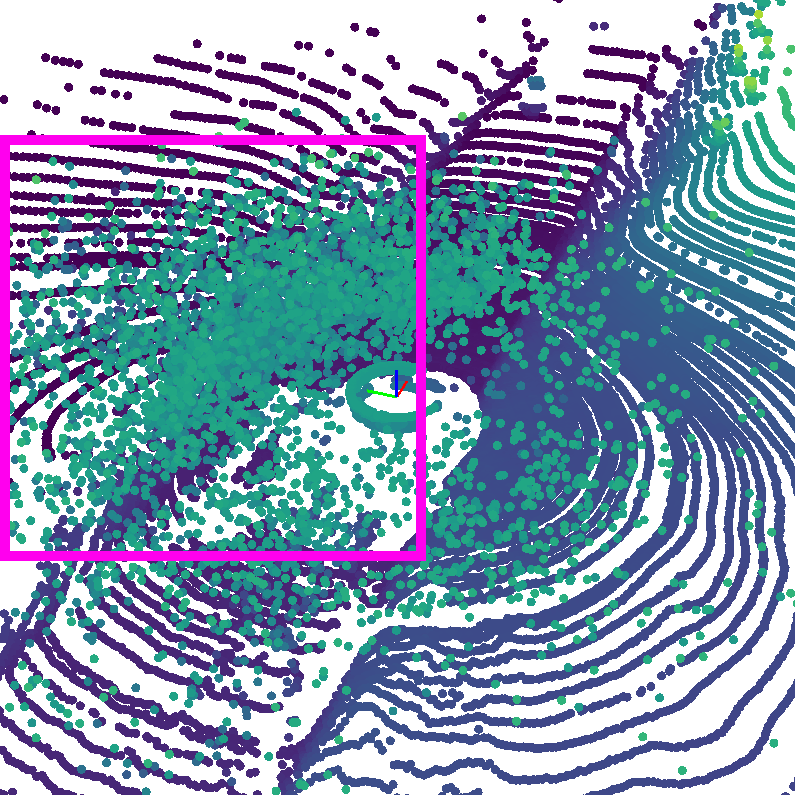}}
         \end{subfigure}
         \begin{subfigure}[b]{0.38\textwidth}
            \frame{\includegraphics[width=0.49\textwidth]{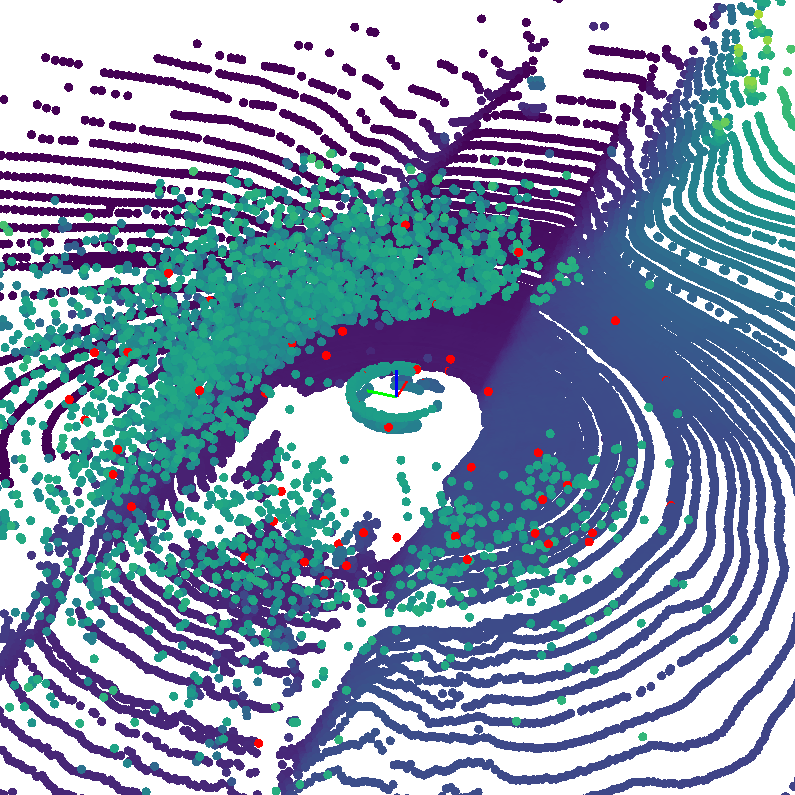}}
            \hspace{-1.5mm}
            \frame{\includegraphics[trim={0 238px 378px 140px}, clip, width=0.46\textwidth, cfbox=Magenta2 1.5pt 1.5pt]{medror_seq4_456.png}}
         \end{subfigure}
         \begin{subfigure}[b]{0.38\textwidth}
            \frame{\includegraphics[width=0.49\textwidth]{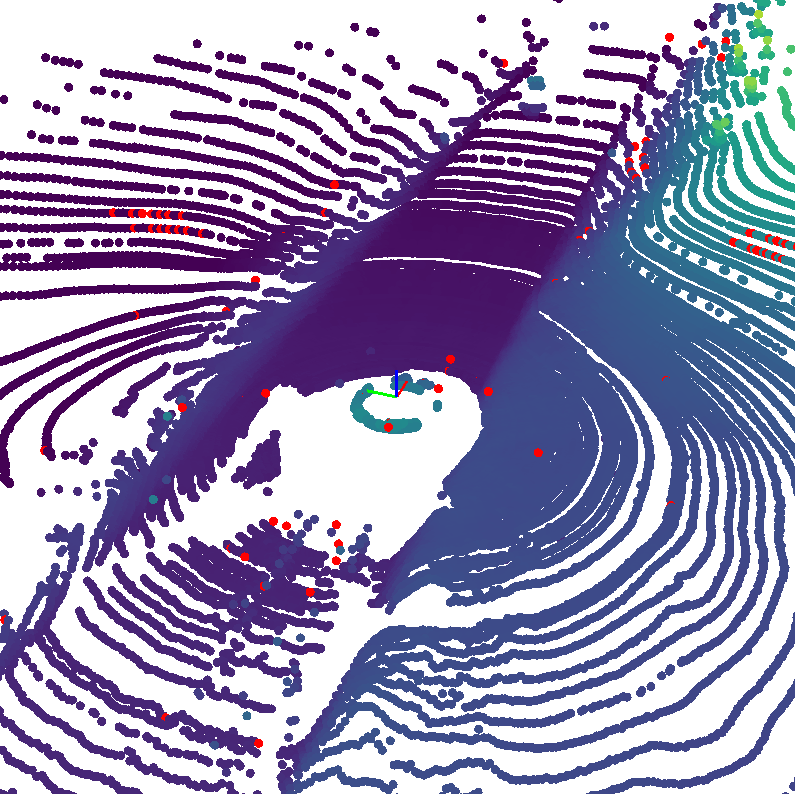}}
            \hspace{-1.5mm}
            \frame{\includegraphics[trim={0 238px 378px 140px}, clip, width=0.46\textwidth, cfbox=Magenta2 1.5pt 1.5pt]{smednet_seq4_456.png}}
         \end{subfigure}
         \vspace{3.0mm}
     \end{subfigure}
     
     \caption{Multi-echo denoising performance on real data. Each row is an individual sample. The corrupted strongest echo is on the leftmost column, the output of the baseline method MEDROR is in the middle, and the output of our SMEDNet is on the right column. \textbf{Red indicates potential substitute points}. MEDROR mostly fails, whereas our SMEDNet picks successfully viable substitute points.}
     \label{fig:multi-echo-results}
\end{figure*}

Multi-echo experiments are evaluated qualitatively as labeled data is not available.
We analyze the performance on the STF \cite{Bijelic_2020_STF} dataset.
The results are presented in Fig. \ref{fig:multi-echo-results}, where each row indicates an individual sample.
The corrupted strongest echo is on the left-hand side, where a detail window is denoted with fuchsia. 
The strongest echo is visualized to highlight the idea of recovering substitute points from alternative echoes.
With that, we want to emphasize that the input to the methods is the multi-echo point cloud.
The baseline method is next to the corrupted strongest echo and our SMEDNet is on the right-hand side.
For these, far-away views and detailed close-ups are visualized.

We compare our method to the baseline Multi-Echo DROR (MEDROR). 
It is a simple modification to the classical DROR \cite{charron2018noising}.
In particular, the neighbors are computed for each echo, where the reference point cloud is the strongest point cloud.
This is similar to the procedure presented in Fig. \ref{fig:knn-multi-echo}.
The searched neighbors are then thresholded similarly to \cite{charron2018noising} but instead of just computing this for the strongest echo it is computed for all echoes in the echo group.
Then, based on the inlier-outlier classification of \cite{charron2018noising}, we classify the points according to Section \ref{sec:inference}.
Note that, the score is now binary instead of a float.
Please refer to the supplementary paper for more details.

The substitute points are visualized in red. 
These are recovered points from alternative echoes (in our case conditional last and second strongest).
As seen in Fig. \ref{fig:multi-echo-results}, SMEDNet successfully finds viable substitute points.
The results prove the concept of multi-echo denoising in adverse weather. 
They also demonstrate that a simple classical method such as MEDROR is not sufficient as there are only a few recovered substitute points and many false negatives.

Classical MEDROR performs much better with low levels of noise compared to a high level of noise.
Note here that the threshold of MEDROR can be adjusted but we noticed that it results in a vast amount of false positives.
Contrarily, our SMEDNet performs well in all levels of noise, even in extreme conditions in Fig. \ref{fig:sample5}.
This shows that our method achieves superior performance.

\noindent \textbf{Limitation.}
As stated in Section \ref{sec:implementation} the number of echoes is limited to the strongest, conditional second strongest, and last echo due to the LiDAR model.
The proposed task and method should be therefore studied with a LiDAR that produces more echoes, for instance, an arbitrary number of strongest echoes and the last echo \textit{e.g.} 1st strongest, 2nd strongest, 3rd strongest, ..., and last echo.
We hypothesize that more echoes are beneficial as the potential substitutes can be in any echo number.
This is a great opportunity for future research, and we hope that LiDAR manufacturers could provide such equipment. 



\section{Conclusion}
\label{sec:conclusion}

We proposed the task of multi-echo denoising in adverse weather.
The idea is to recover points that carry useful information from alternative echoes.
This can be thought of as seeing through the adverse weather-induced noise. 
We also proposed a self-supervised deep learning method for achieving this goal, namely SMEDNet, and proposed characteristics similarity regularization to boost its performance.
Our method achieved new state-of-the-art self-supervised performance on a semi-synthetic dataset. 
Moreover, strong qualitative results on a real multi-echo adverse weather dataset prove the concept of multi-echo denoising.
Compared to a classical MEDROR we show the superiority of deep learning-based SMEDNet.
Based on the results, our work enables safer and more reliable LiDAR point cloud data in adverse weather and therefore should increase the safety of, for instance, autonomous driving and driving assistance systems. 


\bibliographystyle{unsrt}
\bibliography{bibliography}

\end{document}